\documentclass{article}

\usepackage[preprint]{style/neurips_2022}
\usepackage{style/algorithm}
\usepackage{style/algorithmic}

\usepackage[utf8]{inputenc} 
\usepackage[T1]{fontenc}    
\usepackage{hyperref}       
\usepackage{url}            
\usepackage{booktabs}       
\usepackage{amsfonts}       
\usepackage{nicefrac}       
\usepackage{microtype}      
\usepackage{xcolor}         

\usepackage{amsmath}
\usepackage{amssymb}
\usepackage{mathtools}
\usepackage{amsthm}

\theoremstyle{plain}
\newtheorem{theorem}{Theorem}[section]

\theoremstyle{definition}
\newtheorem{definition}[theorem]{Definition}

\theoremstyle{remark}

\usepackage{subcaption}
\usepackage{graphicx}

\title{FIB: A Method for Evaluation of Feature Impact Balance in Multi-Dimensional Data}

\author{
    Xavier F. Cadet\thanks{\texttt{xfc17@ic.ac.uk}} \\ 
    Imperial College London \\ 
    \And
    Sara Ahmadi-Abhari \\
    Imperial College London \\ 
    \And
    Hamed Haddadi \\
    Imperial College London \\ 
}

\begin{document}

\maketitle

\begin{abstract}

Errors might not have the same consequences depending on the task at hand. Nevertheless, there is limited research investigating the impact of imbalance in the contribution of different features in an error vector. Therefore, we propose the Feature Impact Balance (FIB) score. It measures whether there is a balanced impact of features in the discrepancies between two vectors. We designed the FIB score to lie in [0, 1]. Scores close to 0 indicate that a small number of features contribute to most of the error, and scores close to 1 indicate that most features contribute to the error equally.
We experimentally study the FIB on different datasets, using AutoEncoders and Variational AutoEncoders. We show how the feature impact balance varies during training and showcase its usability to support model selection for single output and multi-output tasks.

\end{abstract}

\section{Introduction}
 \label{sec:intro}

In an healthcare-related problem failing to correctly predict a single piece of information could have dramatic consequences. On the other hand, poorly predicting a single pixel in the background of an image could be acceptable. Such discrepancies between the two scenarios highlight that even a single feature can make a difference.

While there are plenty of metrics to evaluate model performance, to the best of our knowledge, no metric is concerned with the balance of features impact on a given error. Therefore, we introduce the Feature Impact Balance (FIB) score, a metric that quantifies to which extent errors between two vectors depend on a small number of features or many of them. In this work, we look into different properties of models and representations and their effect on the FIB score.

In this paper, we attempt to answer the following questions:
(i) How are AutoEncoders equilibrating their errors during training?
(ii) Can we use the FIB score to find the best models?
(iii) Is the representation learned by these better suited for downstream tasks such as classification regression?

To answer these questions we focus on a specific type of models, namely AutoEncoders (AE) and Variational AutoEncoders(VAE) \cite{kingmaAutoEncodingVariationalBayes2014}
commonly used for unsupervised learning task. We use shallow AutoEncoders and relatively low dimensional datasets for visualization purposes and to help build intuition.
We look into learning and re-purposing a representation for a given task to another task, commonly considered as Transfer Learning problem. Transfer Learning helped advance numerous fields such as drug discovery \cite{ramsundarMassivelyMultitaskNetworks2015, yangLongitudinalPredictiveModeling2021, minProteinTransferLearning2021} or natural language processing \cite{liuMultiTaskDeepNeural2019, leePredictingWhatYou2021}.
One approach to Transfer Learning relies on the assumption that a common representation exists suitable for multiple tasks \cite{duFewShotLearningLearning2021, tripuraneniProvableMetaLearningLinear2021}. We also assume that there exists a representation shareable across multiple tasks.
  
The contributions of this paper are:
\begin{itemize}
    \item \textit{Feature Impact Balance (FIB) score} a mathematically-grounded quantification of the balance of the contribution of each feature in the difference between two vectors.
    \item Empirical study of the learning pattern and feature balance during training of specific AutoEncoders architectures.
\end{itemize}
\section{Background}
    Here we introduce the problem and illustrate it through a comparison to Mean Squared Error (MSE), we then cover AutoEncoders, and Representation learning.
    
    \paragraph{Problem setting.}
    When comparing multidimensional continuous outputs, various methods exist based on the data type.
    A common method to quantify dissimilarity between two continuous multidimensional objects is the Mean Squared Error (MSE).
    When two vectors $\mathbf{x}$ and $\mathbf{y}$ represent multiple observations of their respective variables, the MSE quantifies the average difference between each observation $x_i$ and $y_i$. 
    Nonetheless, if we consider these vectors as single observations with multiple properties, the MSE quantifies, on average, how each observation's properties are dissimilar from one another. In this paper, we consider the MSE as the latter version.

\begin{figure}[ht]
        \vskip 0.2in
            \begin{center}
                \centerline{\includegraphics[width=\columnwidth/2]{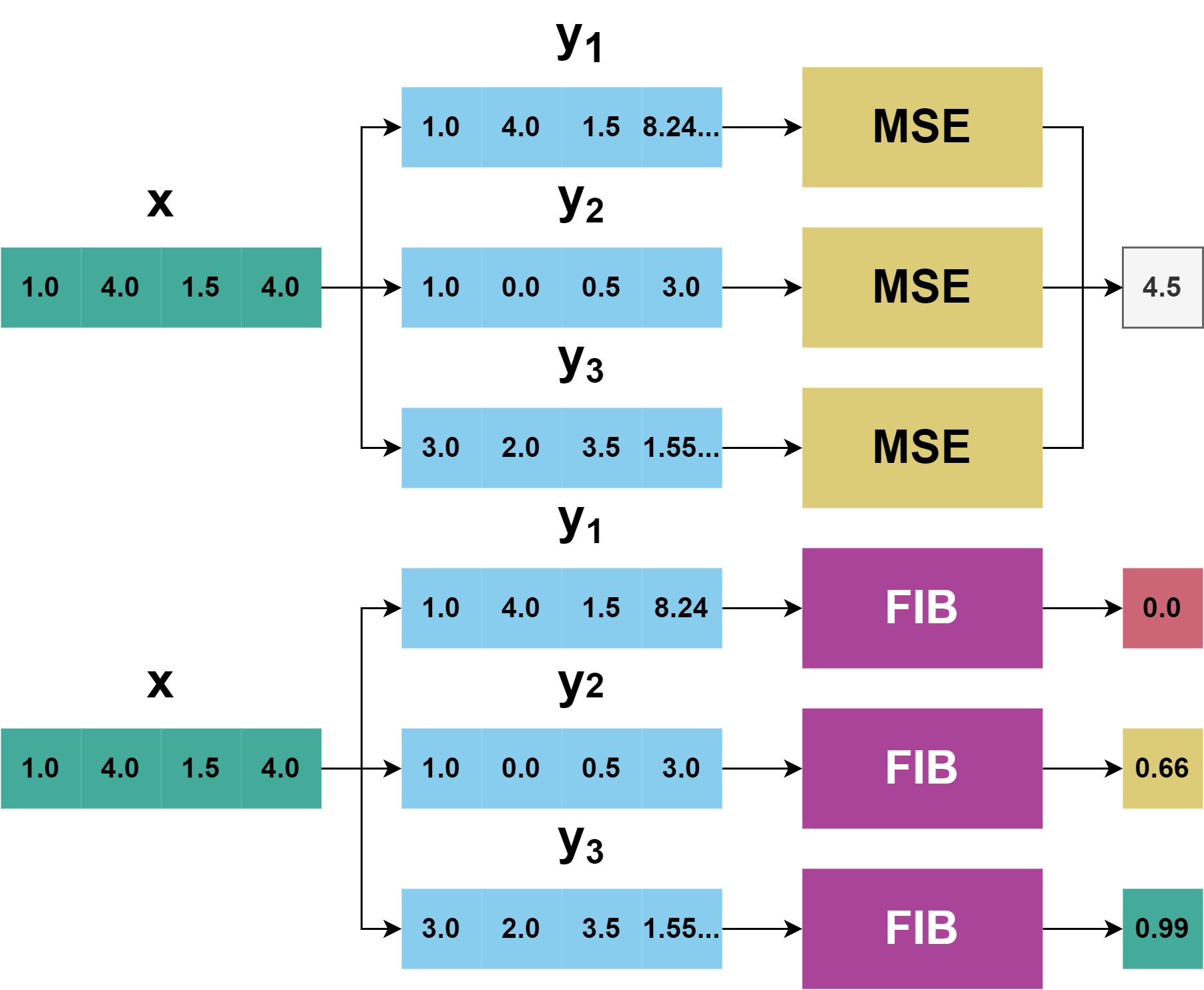}}
                 \caption{Comparison between MSE and FIB: Vectors $\mathbf{x}$ represent the ground truth, vectors $\mathbf{y}_1, \mathbf{y}_2, \mathbf{y}_3$ represent hypothetical predictions from models. The errors between $\mathbf{x}$ and the $\mathbf{y}$s are carried by different columns.
                 The MSE yields the same value for all combinations of $\mathbf{x}$ and $\mathbf{y}$s. The FIB indicates (from top to bottom) an important imbalance when looking at $\mathbf{y}_1$, moderated imbalance using $\mathbf{y}_2$, and balanced error contribution considering $\mathbf{y}_3$.}
                \label{figure:mse_vs_fib}
            \end{center}
        \vskip -0.2in
    \end{figure}

    One of the shortcomings of the MSE is that it can associate the same value to vectors with drastically different feature impacts as illustrated in Figure \ref{figure:mse_vs_fib}. The MSE does not distinguish the cases where a single feature, a subgroup of features, or all the features, carry the errors.
    
    \paragraph{Representation Learning.} 
    Over the past years, the interest in Representation Learning has grown, becoming a field on its own \cite{bengioRepresentationLearningReview2014}. Representation Learning aims at finding a description of data, a new representation, which makes subsequent tasks simpler to solve \cite{goodfellowDeepLearning2016}.
    A common trade-off between preserving information about the input and obtaining properties of interest exists in Representation Learning problems \cite{goodfellowDeepLearning2016}.
 
    \paragraph{AutoEncoder.} 
    AutoEncoders are Neural Network architectures that are extensively used for Unsupervised Machine Learning tasks. Their aim is to learn a new representation of their input, such that they can reproduce their inputs \cite{rumelhartLearningInternal1987, baldiNeuralNetworksPrincipal1989, hintonAutoencodersMinimumDescription1993, baldiAutoencodersUnsupervisedLearning2011}.
    AutoEncoders are composed of 2 sections: an encoder function $f$ and a decoder function $g$. The encoder maps the input $\mathbf{x}$ from the input space to the code space resulting in a code $\mathbf{h}$. The decoder maps the code $\mathbf{h}$ from the code space to the output space and produces $\hat{\mathbf{x}}$, which stands for the reconstruction of $\mathbf{x}$.
\section{Feature Impact Balance (FIB)}
\label{sec:fib}
    Let's compare a vector $\mathbf{x}$ to a group of vectors $\{\mathbf{y}_1, \mathbf{y}_2, \mathbf{y}_3\}$.
    We might obtain the same MSE values even if the dissimilarities per feature are drastically different, as illustrated in Figure \ref{figure:mse_vs_fib}. Having errors evenly spread across all features can be beneficial if we do not know which feature is the most important. On the contrary, if we know that a feature is important, we should pick a model that does not fail to predict this feature.
    
    To quantify the balance between the contributions of each feature in the error, we introduce the Feature Impact Balance (FIB) score. In this section, we define the different components leading to the computation of the FIB, from the Internal Error (IE) quantification, the Feature Impact (FI), the Feature Impact Imbalance (FII), and the proofs that guarantee that the FIB score can take values in $[0, 1]$.
    Having values ranging in $[0, 1]$ allows comparing models across experiments, similarly to the accuracy score in classification tasks. 
    
    We define the FIB in its simplest form as :
    \begin{equation}
        FIB(\mathbf{x}, \mathbf{y}) = 1 - \frac{M}{M - 1}\sum_{k=1}^{M}(\frac{|x_{k} - y_{k}|}{||\mathbf{x} - \mathbf{y}||_{1}} - \frac{1}{M})^2 \in [0, 1]
        \label{equation:feature_impact_balance_complete}
    \end{equation}
    
    With $\mathbf{x}, \mathbf{y} \in \mathbb{R}^{M}$ and $|| \mathbf{x} - \mathbf{y}||_1 = \sum_{k=1}^{M}|x_k - y_k|$
    
    The Feature Impact Balance quantifies errors at two levels, the Internal Error (IE) and Balance Error (BE).
    The Internal Error quantifies the error between the two entries $\mathbf{x}$ and $\mathbf{y}$. The Balance Error quantifies the error between the contribution of each entries in the Internal error and a balance vector. The balance vector represents the scenario where each component contributed equally to the error.
    
    In the following, we detail the steps to compute the FIB score between two vectors; the computations can be adapted to matrices. Using matrices, we evaluate properties at different levels. For instance, we could consider subsets of a given dataset, features associated with a specific group, or even mini-batch when training a Neural Network.
    
    Let $\mathbf{x}, \mathbf{y} \in \mathbb{R}^M$ where $M$ is the number of features. We define the Internal Error (IE) as a function that takes two entries $\mathbf{x}$ and $\mathbf{y}$ and results in a vector $\mathbf{e} \in \mathbb{R}^K$ ($e_{k}$ indicates the $k$-th element of the vector $\mathbf{e}$):
    \begin{equation}
        IE(\mathbf{x}, \mathbf{y}) = \mathbf{e} \in \mathbb{R}^{K}
        \label{equation:internal_error_function}
    \end{equation}
    where $K$ is the dimension of the error vector. We allow the dimension of the Internal Error to be different from those of the inputs, such that we can consider feature groups in the next sections.
    
    In the following steps, we consider the Absolute Error as Internal Error function, yielding a vector of the same dimensions as the inputs (\textit{i.e} $K=M$). We expect the Internal Error function to be constituted of positive or null values (null values when no error is made). Therefore, we use:
    \begin{equation}
        IE(\mathbf{x}, \mathbf{y}) = AE(\mathbf{x}, \mathbf{y})
    \label{equation:interal_error_ae}
        = [|x_1 - y_1|, ..., |x_M - y_M|]^T \in \mathbb{R}^{M}
    \end{equation}
    \begin{equation}
        AE(\mathbf{x}, \mathbf{y})= [|x_1 - y_1|, ..., |x_M - y_M|]^T \in \mathbb{R}^{M}
    \label{equation:what_is_ae}
    \end{equation}
    
    We then define the Feature Impact (FI) as the Internal Error vector divided by the sum of its elements:
    \begin{equation}
        FI(\mathbf{x}, \mathbf{y}) = \frac{\mathbf{e}}{\sum_{k=1}^{M}e_k} \in [0, 1]^M
        \label{equation:feature_impact_vector}
    \end{equation}
    where $e = IE(\mathbf{x}, \mathbf{y})$. It follows that the sum of its elements equals to $1$. 
    
    The elements of the Feature Impact vector indicates the contribution of each feature to the errors. 
    
    In case there are no errors, (the errors between $\mathbf{x}$ and $\mathbf{y}$ sum to $0$), we substitute $FI(\mathbf{x}, \mathbf{y})$ by $\frac{\mathbf{1}_M}{M}$
    where $\mathbf{1_M} = [1, ..., 1]^T \in \mathbb{R}^M$. 
    
    As such, we consider that when no errors are done, each component contributed equally.
    
    To quantify the imbalance of the contributions of each feature, we define the Balance Error (BE). When each feature participates equally (namely contributes to $\frac{1}{M}$ of the sum of the errors), the BE equals $0$. Otherwise, it quantifies how far the contributions are from the expected contributions.
    
    Then, we define the Feature Impact Imbalance (FII) as the Balance Error between the Feature Impact vector from equation \ref{equation:feature_impact_vector} and a Balance vector.
    The Balance vector is $\mathbf{b} = \frac{\mathbf{1}_M}{M} \in \mathbb{R}^M$.
    Therefore, we define the FII as:
    \begin{equation}
        FII(\mathbf{x}, \mathbf{y}) = BE(FI(\mathbf{x}, \mathbf{y}), \frac{\mathbf{1}_M}{M})
        \label{equation:feature_impact_imbalance}
    \end{equation}
    
    We use the MSE to compute the Balance Error in the following experiments. Nonetheless, we could also use the Mean Absolute Error (MAE).
    
    By computing the Squared Error between the Feature Impact vector and the balance vector, we evaluate the error between the Feature Impact observed and the Feature Impact achieved when every feature contributed equally.
    Finally, we take the mean of these balance errors to obtain a single value to quantify the imbalance.

    The shortcoming from the MSE persists here, but at a different level, the FII will not distinguish between different combinations
    of feature impact imbalances (i.e $[1.0, 0.0]^T$ against $[0.0, 1.0]^T$ in $\mathbb{R}^2$), but it will be sensible to the differences at the input level $\mathbf{x}$ and $\mathbf{y}$.
    
    We demonstrate that when using the MAE and MSE as Balance error, the FII can reach a maximum via Theorem \ref{theorem:reach_max} and its value using Theorem \ref{theorem:maximal_value}.
   
    Let $(c_k)_{k \in \{1, \cdot \cdot \cdot M\}}$ be the values of $FI$ such  that $FI(\mathbf{x}, \mathbf{y}) = [c_1, \cdot \cdot \cdot, c_M]^T$. By construction $\forall k \in \{1, \cdot \cdot \cdot, M\}, 0 \leq c_k \leq 1$ and $\sum_{k=1}^{M}c_k = 1$.
    The right hand part of the equation \ref{equation:feature_impact_imbalance} when using the MAE can be  rewritten as a function of $M$ variables noted $\varphi(c_1, ..., c_M): \mathbb{R}^{M} \rightarrow \mathbb{R}$
    \begin{equation}
        \varphi(c_1, ..., c_M) = \frac{1}{M}\sum_{k = 1}^{M}|c_k - \frac{1}{M}|
        \label{equation:function_fii_using_mae}
    \end{equation}
    
    Similarly when using the MSE, we now consider $\psi(c_1, ..., c_M): \mathbb{R}^{M} \rightarrow \mathbb{R}$
    
    \begin{equation}
        \psi(c_1, ..., c_M) = \frac{1}{M}\sum_{k = 1}^{M}(c_k - \frac{1}{M})^2
        \label{equation:function_fii_using_mse}
    \end{equation}
 
    In order to compute the maximum of $\varphi$ and $\psi$, we first give the
    following definition.
    
    \begin{definition}
    \cite{zieglerLecturesPolytopes1995}
    A $M$-simplex in $\mathbb{R}^{M}$, $M\ge 1$, is the convex
    hull of $M+1$ affinely independent points.
    \end{definition}
    A $M$-simplex is the intersection of all convex sets containing these points; it is also the smallest convex set containing these points. It finally follows from a theorem due to Carath\'eodory \cite{zieglerLecturesPolytopes1995} that the $M$-simplex $S$ defined by the points $a_1$, $\cdot \cdot \cdot $, $a_{M+1}$ is characterized as follows: a point $a$ belongs to $S$ if and only if there exists $\lambda _i$, $i=1$, $\cdot \cdot \cdot $, $M+1$, $\lambda _i\in [0,1]$, $i=1$, $\cdot \cdot \cdot $, $M+1$, $\sum _{i=1}^{M+1}\lambda _i=1$, such that
    
    $$a=\sum _{i=1}^{M+1}\lambda _ia_i.$$
    
    This yields that a $M$-simplex is closed and bounded, and thus compact, in $\mathbb{R}^{M}$; it is also, by definition, convex.
    \begin{theorem}
        \label{theorem:reach_max}
        Let $f:S \rightarrow \mathbb{R}$ be continuous and convex, where $S \subset \mathbb{R}^{M}$ is the $M$-simplex defined by the points $a_1$, $\cdot \cdot \cdot $, $a_{M+1}$, $M\ge 1$. Then $f$ reaches its maximum at one of the $a_i$'s.

    \end{theorem}
    
    \begin{theorem}
        \label{theorem:maximal_value}
        Let $A = \{(c_1, ..., c_M) \in \mathbb{R}^M, c_i \in [0, 1], i = 1, .., M, \sum_{i=1}^{M}c_i = 1\}$. We set
$\varphi (c_1, ..., c_M) = \sum_{i= 1}^{M}|c_i - \frac{1}{M}|$ and $\psi (c_1, ..., c_M) = \sum_{i = 1}^{M}(c_i - \frac{1}{M})^2$. Then, $\max_{A} \varphi = 2\frac{ M -1}{M}$ and $\max_{A} \psi = \frac{M - 1}{M}$.
    \end{theorem}
    
    We provide the proofs for both theorems in the Appendix \ref{appendix:theorems_and_proofs}.
    When using the MSE as Balance Error, we know from Theorem \ref{theorem:reach_max} that: 
    \begin{equation}
        0 \leq FII(\mathbf{x}, \mathbf{y}) \leq \frac{M - 1}{M^2}
        \label{equation:fii_mse_boundaries}
    \end{equation}
     where the value $0$ means that the feature impact is balanced, as it is achieved in the case where $\forall k \in 1, ..., M, c_k = \frac{1}{M}$.
     
    Using the maximal values of the FII, we can now normalize the FII scores and obtain values in $[0, 1]$.
    For a more straightforward interpretation of the FII values, we define the Normalized Feature Impact Imbalance (NFII) as:
    \begin{equation}
        NFII(\mathbf{x}, \mathbf{y}) = \frac{M^2}{M - 1}FII(\mathbf{x}, \mathbf{y})
        \label{equation:nfii}
    \end{equation}
    From equations \ref{equation:fii_mse_boundaries} and \ref{equation:nfii} it comes that:
    \begin{equation}
        0 \leq \frac{M^2}{M - 1}FII(\mathbf{x}, \mathbf{y}) \leq 1
        \label{equation:boudnaries_nfii}
    \end{equation}

    Finally, we introduce the Feature Impact Balance (FIB). For conciseness, we omit the mention that it is normalized, as one minus the NFII, defined as:
    \begin{equation}
        FIB(\mathbf{x}, \mathbf{y}) = 1 - NFII(\mathbf{x}, \mathbf{y}) \in [0, 1]
        \label{equation:feature_impact_balance}
    \end{equation}
    
    A value of 1 indicates that each of the features contributed equally to the errors, and a value of 0 indicates that a single feature is responsible for the error.

    \subsection{Adaptation to higher number of features.}
        \label{sec:exp_higher_num_features}
        The more features there are, the higher the chances that the FIB score will be high. To reach a balanced impact each feature needs to contribute to $\frac{1}{M}$. Therefore, to obtain a score of 0, it requires that a single feature carries all the error. Furthermore, as the number of features increases the contribution required for balance gets close to 0.
        
        In order to use the FIB on high dimensional data, we propose to group features. For instance, considering an image dataset such as MNIST with images composed of 28x28 pixels, namely 784 features, we can regroup these features into 10 groups, and compute the FIB score over these 10 groups instead of the 784 features.

\begin{figure}[h]
            \centering
            \begin{subfigure}{0.32\textwidth}
                \centering
                \includegraphics[width=\textwidth]{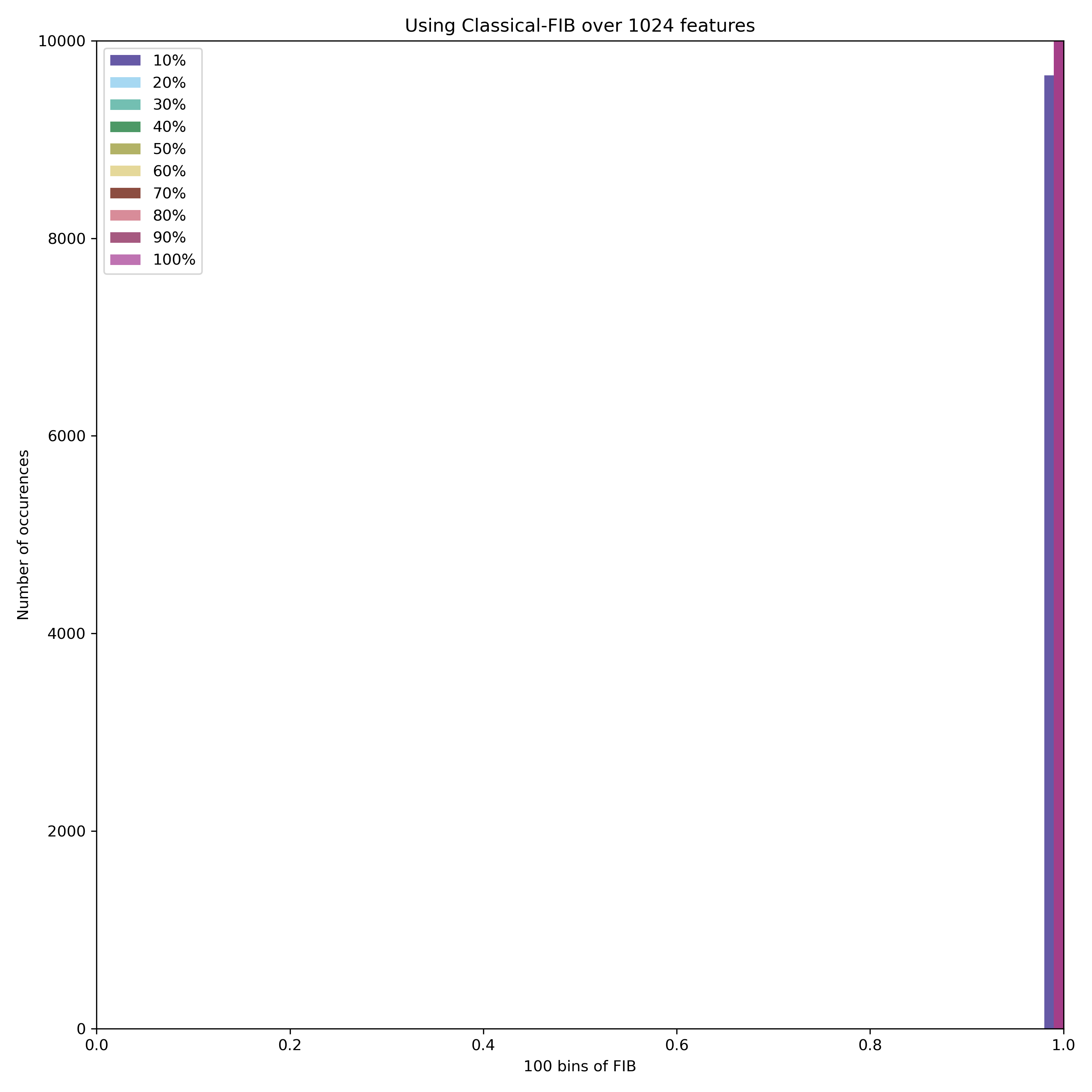}
                \caption{No groups}
                \label{fig:sub_grouping_density_no_groups}
            \end{subfigure}
                \begin{subfigure}{0.32\textwidth}
                \centering
                \includegraphics[width=\textwidth]{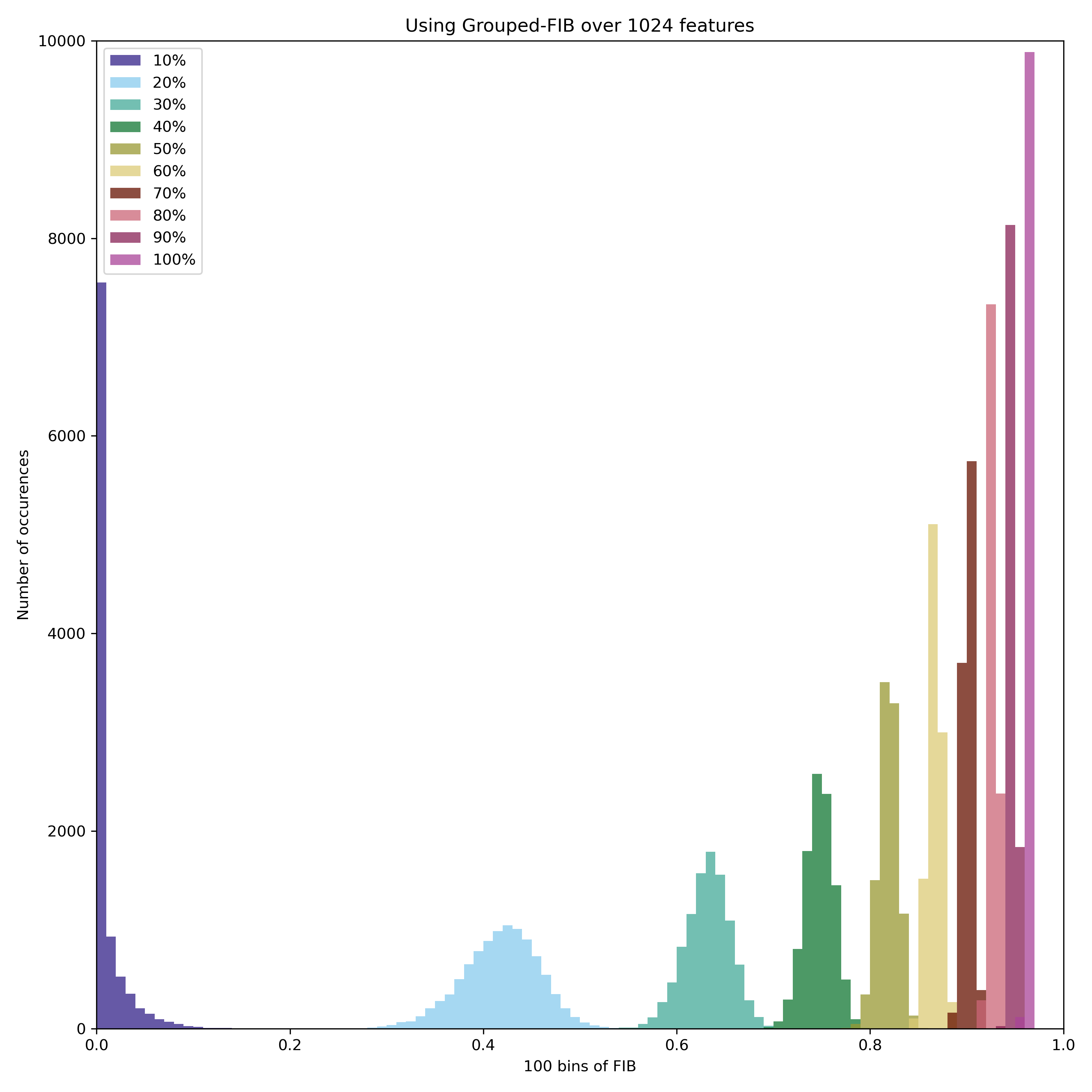}
                \caption{10 Groups}
                \label{fig:sub_grouping_density_10_groups}
                \end{subfigure}
            \caption{We add noise drawn from an Uniform distribution into a 1024-Dimensional vector for different proportions of perturbation (10\% to 100\%). We repeat this operation 10000 times. In \ref{fig:sub_grouping_density_no_groups} we compute the FIB without grouping. In \ref{fig:sub_grouping_density_10_groups} we compute the FIB over 10 groups. While for the FIB without groups most scores are close to 1.0, for the 10 Groups FIB, the scores are more spread over [0, 1].}
            \label{fig:feature_grouping_1024}
        \end{figure}

        In Figure \ref{fig:feature_grouping_1024} we give an example with $M=1024$, comparing using no groups (Figure \ref{fig:sub_grouping_density_no_groups}) and $10$ groups (Figure \ref{fig:sub_grouping_density_10_groups}). After injecting noise in 10\% of the 1024 features, we obtain a score that is already close to $1.0$, \textit{i.e} $> 0.99$ (99\% in the plot). Nonetheless, after modifying a single feature, the score drops to $0$. 
        While it can be an interesting property, for instance when each feature is critical, we might be interested in imbalance between groups of features. For instance, we could consider that anywhere between 1 feature and 10\% of the features should result in a FIB score of 0.0 (0\% in Figure \ref{fig:sub_grouping_density_10_groups}).
        To do so, we can specify a different Internal Error as defined in equation \ref{equation:internal_error_function}. For instance, we can group the $M$ features into $K$ groups. Then, we chose an aggregation strategy (sorting, random selection, predefined groups, etc...). Finally, we pick a reduction strategy to bring each group to a single value (similarly to Mean Pooling or Max Pooling in Convolutional Neural Networks).
        
        We provide an example of grouping using sorting and splitting as an aggregation method and summation as the reduction operation in Appendix \ref{appendix:feature_grouping_algorithm}
\section{Experiments}
    \label{sec:experiments}
    We ran experiments over different datasets (Section \ref{sec:exp_datasets}), we show the impact of Feature Grouping (Section \ref{sec:exp_feature_grouping}) using different types of AutoEncoders (Section \ref{sec:exp_aes}), and show how FIB can be used to support model selection (Section \ref{sec:exp_model_selection}).
    \subsection{Experimental Setup}
        \paragraph{Datasets.}
        We chose three datasets to run experiments on: 1)  Iris \cite{fisherUseMultipleMeasurements1936}, 2) SARCOS \cite{vijayakumarLocallyWeightedProjection2000} and 3) MNIST \cite{lecunMnistDatabaseHandwritten2005}.

       \paragraph{Feature Grouping.}
        We computed grouped FIB scores on SARCOS and MNIST for group number ranging from 2 to 10.

        \paragraph{AutoEncoders.}
        We experiment with two types of AutoEncoders. These are: Classical AutoEncoders (AE) and Variational AutoEncoders (VAE). In both cases, we considered only AutoEncoders based on Fully Connected layers.
        
       \paragraph{Using FIB to support Model Selection.}
        We chose the best performing epoch of the classical AutoEncoders based on their validation loss. We use these models as feature extractors. We then train Logistic Regressions and Linear regression using these features and evaluate their performance on a test set.
            
   \subsection{Different Datasets}
    \label{sec:exp_datasets}
\begin{figure}[h]
        \begin{subfigure}{0.32\textwidth}
            \centering
            \includegraphics[width=\textwidth]{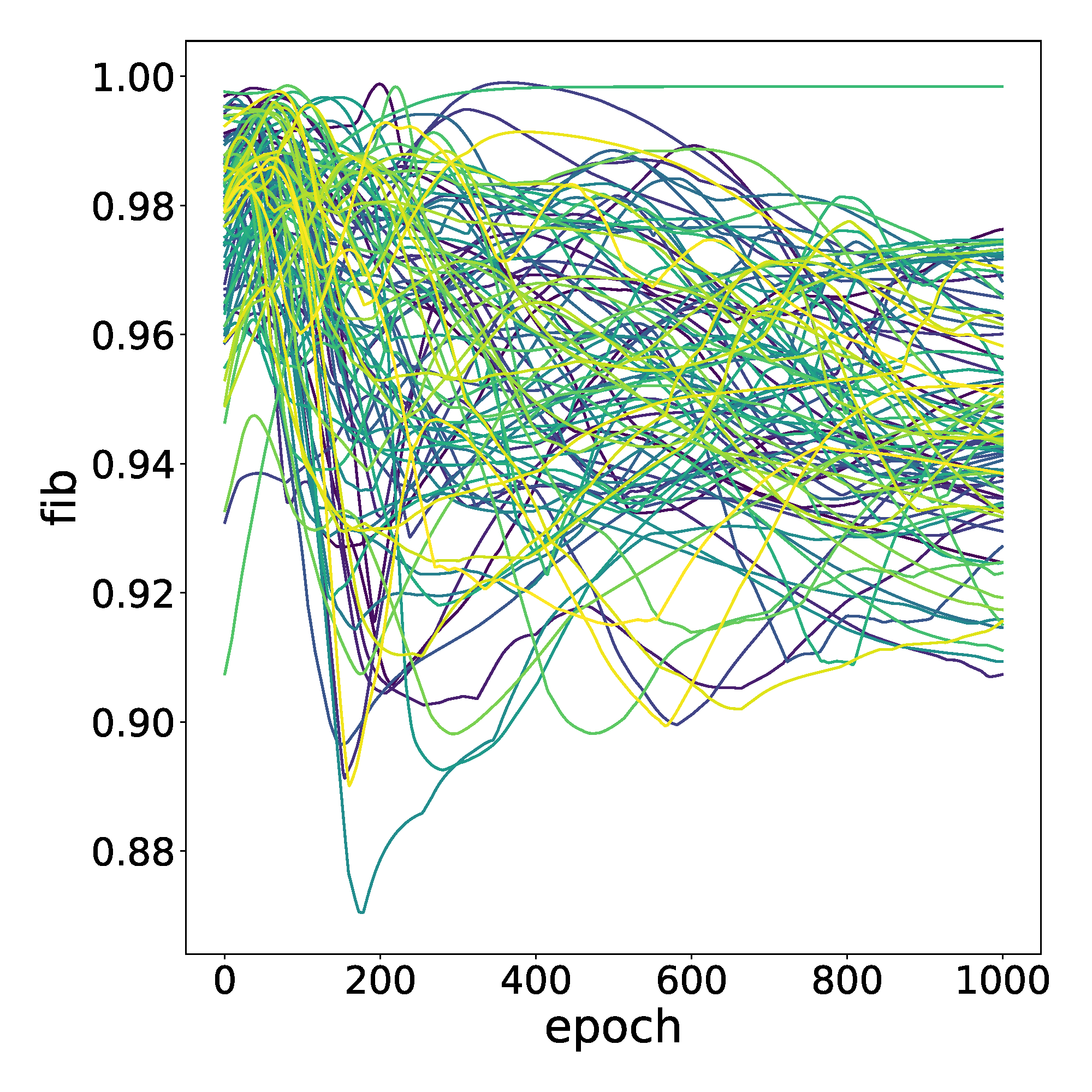}
            \caption{Iris}
            \label{fig:data_iris_ae_val_fib}
        \end{subfigure}
        \begin{subfigure}{0.32\textwidth}
            \centering
            \includegraphics[width=\textwidth]{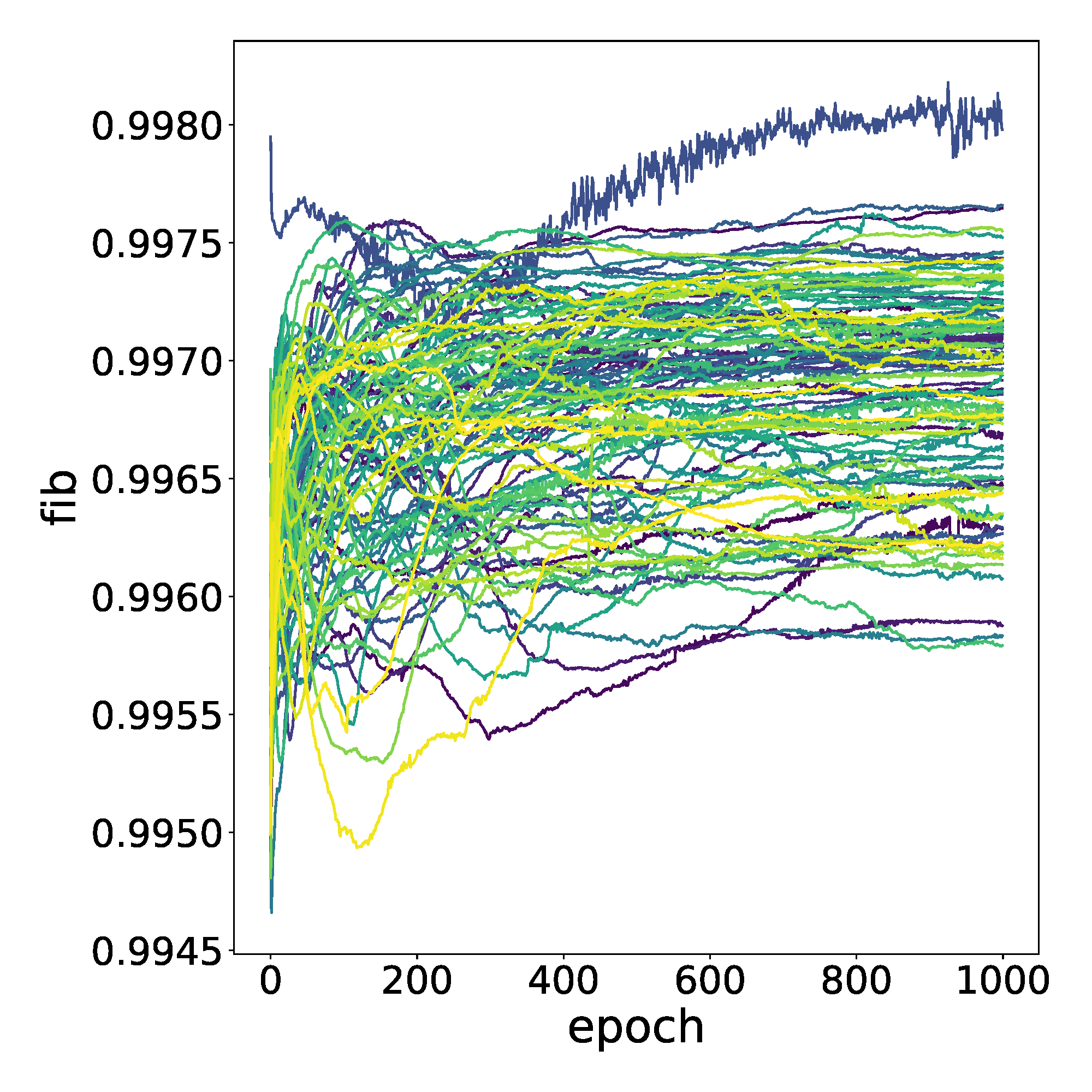}
            \caption{SARCOS}
            \label{fig:data_sarcos_ae_val_fib}
        \end{subfigure}
        \begin{subfigure}{0.32\textwidth}
            \centering
            \includegraphics[width=\textwidth]{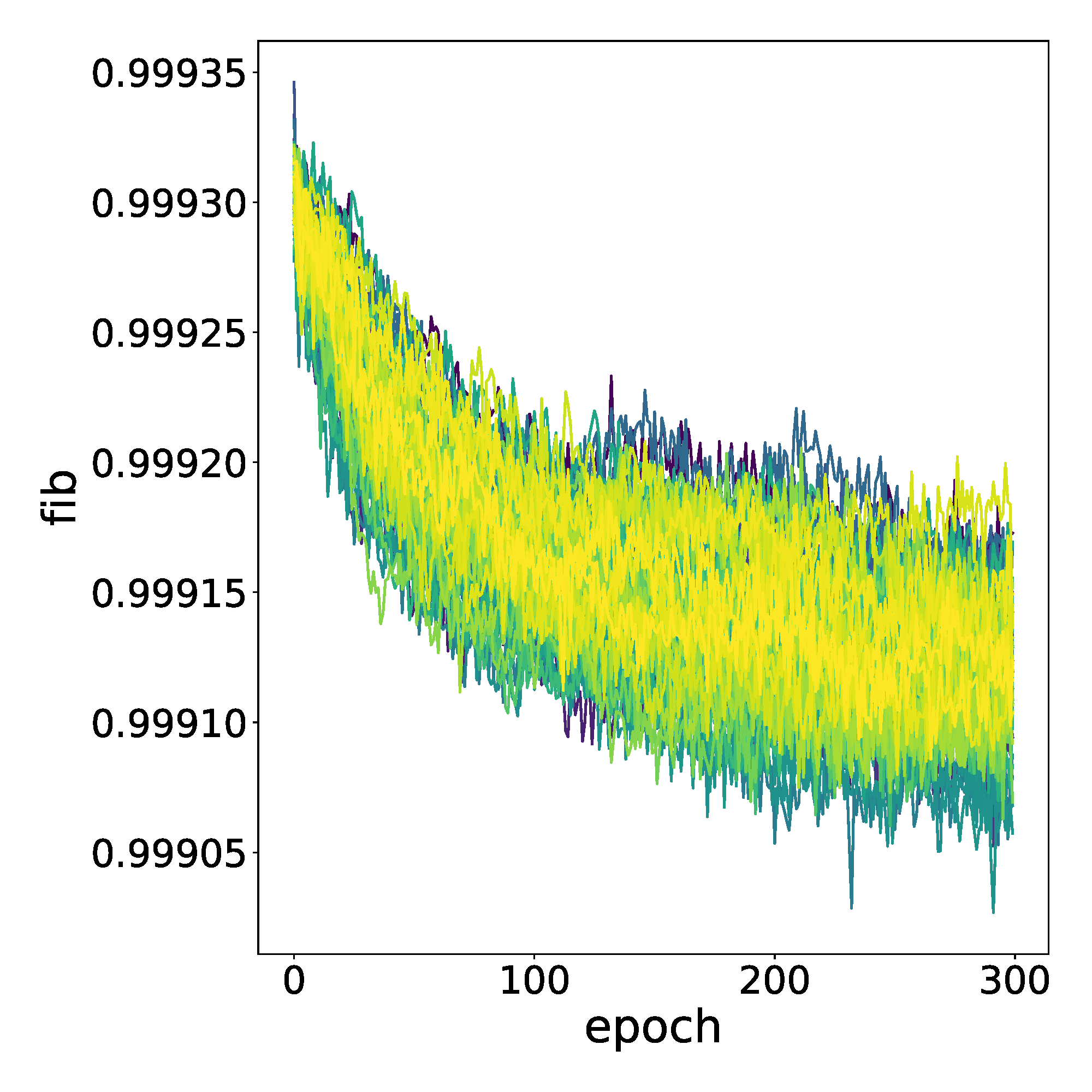}
            \caption{MNIST}
            \label{fig:data_mnist_ae_val_fib}
        \end{subfigure}
    \caption{We compare the evolution of the FIB scores for 100 models over 1000 epochs on the Iris and SARCOS datasets, and 300 epochs on the MNIST dataset.
    } 
    \label{fig:iris_vs_sarcos_vs_mnist_val_fib}
    \end{figure}

    We computed the FIB score on the Iris, SARCOS, MNIST datasets. For each of the Dataset, we tried Fully Connected AutoEncoders. Figure \ref{fig:iris_vs_sarcos_vs_mnist_val_fib} shows the evolution of the FIB scores overs 100 models for 1000 epochs for Iris and SARCOS and 300 for MNIST. Result are reported based on the validation set performances. Figure \ref{fig:data_iris_ae_val_fib} shows the evolution using an AutoEncoder with layers of size 4, 2 on Iris. Figure \ref{fig:data_sarcos_ae_val_fib} shows the evolution with sizes 21, 16, 4 for SARCOS. And Figure \ref{fig:data_mnist_ae_val_fib} shows the evolution for sizes 784, 512, 256 for MNIST.  In the case of Iris, the FIB scores vary between 0.80 and 1.0 as the models train. The pattern highlights that while the majority of the models started with high FIB score on the validation as the model trains, the balance is not maintained. Table \ref{tab:mse_and_fib_irs_sarcos_mnist} highlights that after training, models tend to reach FIB scores above 0.90. As the capacity of the models increase, the FIB score is higher.
    
    \begin{table}
        \caption{MSE and FIB scores obtained at the final epoch over two model designs for Iris, SARCOS and MNIST}
        \label{tab:mse_and_fib_irs_sarcos_mnist}
        \centering
        \begin{tabular}{ c  c c  c c  c c } 
        \toprule
         Dataset & Iris & Iris & SARCOS & SARCOS & MNIST & MNIST \\
         Layers & [4, 1] & [4, 2] & [21, 16, 4] & [21, 16, 8] & [784, 512, 128] & [784, 512, 256] \\
         \midrule
         MSE & $0.344 \pm 0.170$ & $0.151 \pm 0.102$ & $0.168 \pm 0.041$ & $0.068 \pm 0.006$ & $0.004 \pm 0.004$ & $0.003 \pm 0.000$ \\ 
         FIB & $0.934 \pm 0.022$ & $0.947 \pm 0.018$ & $0.997 \pm 0.000$ & $0.998 \pm 0.000$ & $0.999 \pm 0.004$ & $0.999 \pm 0.000$ \\
             \bottomrule
        \end{tabular}
   \end{table}

   \subsection{Feature Grouping}
    \label{sec:exp_feature_grouping}
    Table \ref{tab:feature_grouping} contains the FIB scores using Feature Grouping on SARCOS
    with an AutoEncoder with layers 21, 16, 4 and, MNIST with layers 784, 512, 256. Results are averaged over 100 models after 1000 epochs. We grouped the features using sorting and splitting as aggregation and averaging as reduction.
    We computed grouped FIB score for group sizes 2 to 10.
    For the SARCOS Dataset, moving from 21 features to 2 groups of 10 features yields FIB score that are close to 1.0 with a drop of 0.032 in the FIB score. Nevertheless, on MNIST dataset, using 2 groups of 392 features leads to drop of FIB score by 0.527 score, which indicates that close to 50\% of the features contribute to most of the error.
    \begin{table}
        \caption{FIB computed over different numbers of groups on SARCOS and MNIST. A number of group equal to $M$ (the number of features) is the same as computing the FIB without grouping.}
        \label{tab:feature_grouping}
        \centering
        \begin{tabular}{c c c c c c c} 
            \toprule
             \#Groups & 2 & 3 &  5 & 7 & 10 & $M$ \\
             \midrule
              SARCOS & $0.965 \pm 0.006$ & $0.975 \pm 0.004$ & $0.988 \pm 0.002$ & $0.990 \pm 0.001$ & $0.994 \pm 0.001$ & $0.997 \pm 0.000$\\
              MNIST & $0.472 \pm 0.011$ & $0.697 \pm 0.008$ & $0.839 \pm 0.004$ &  $0.889 \pm 0.003$ & $0.926 \pm 0.002$ & $0.999 \pm 0.000$\\
             \bottomrule
       \end{tabular}
    \end{table}

    \subsection{Different AutoEncoders}
    \label{sec:exp_aes}
\begin{figure}[h]
        \centering
        \begin{subfigure}{0.32\textwidth}
            \centering
            \includegraphics[width=\textwidth]{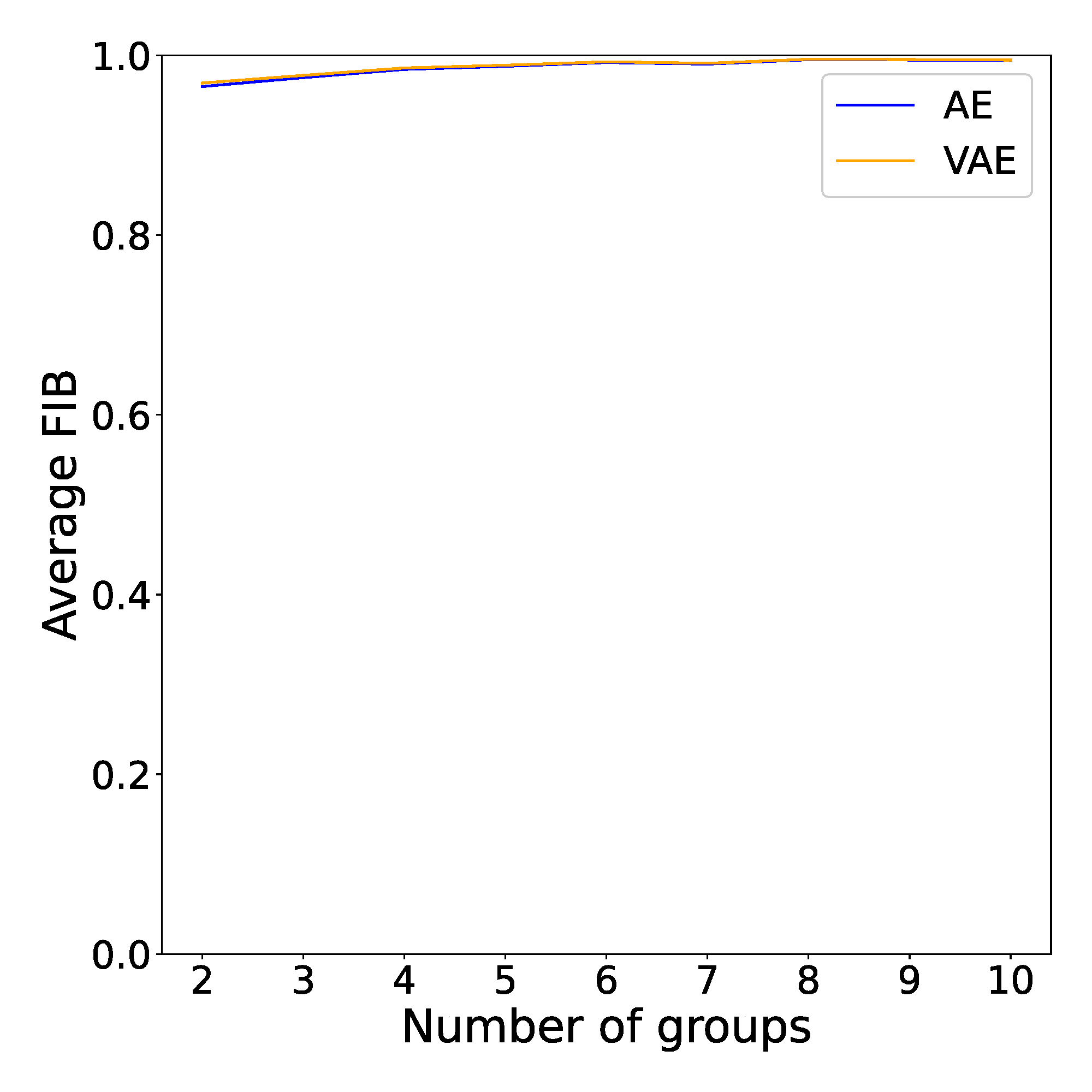}
            \caption{SARCOS}
            \label{fig:exp_vae_sarcos}
        \end{subfigure}
            \begin{subfigure}{0.32\textwidth}
            \centering
            \includegraphics[width=\textwidth]{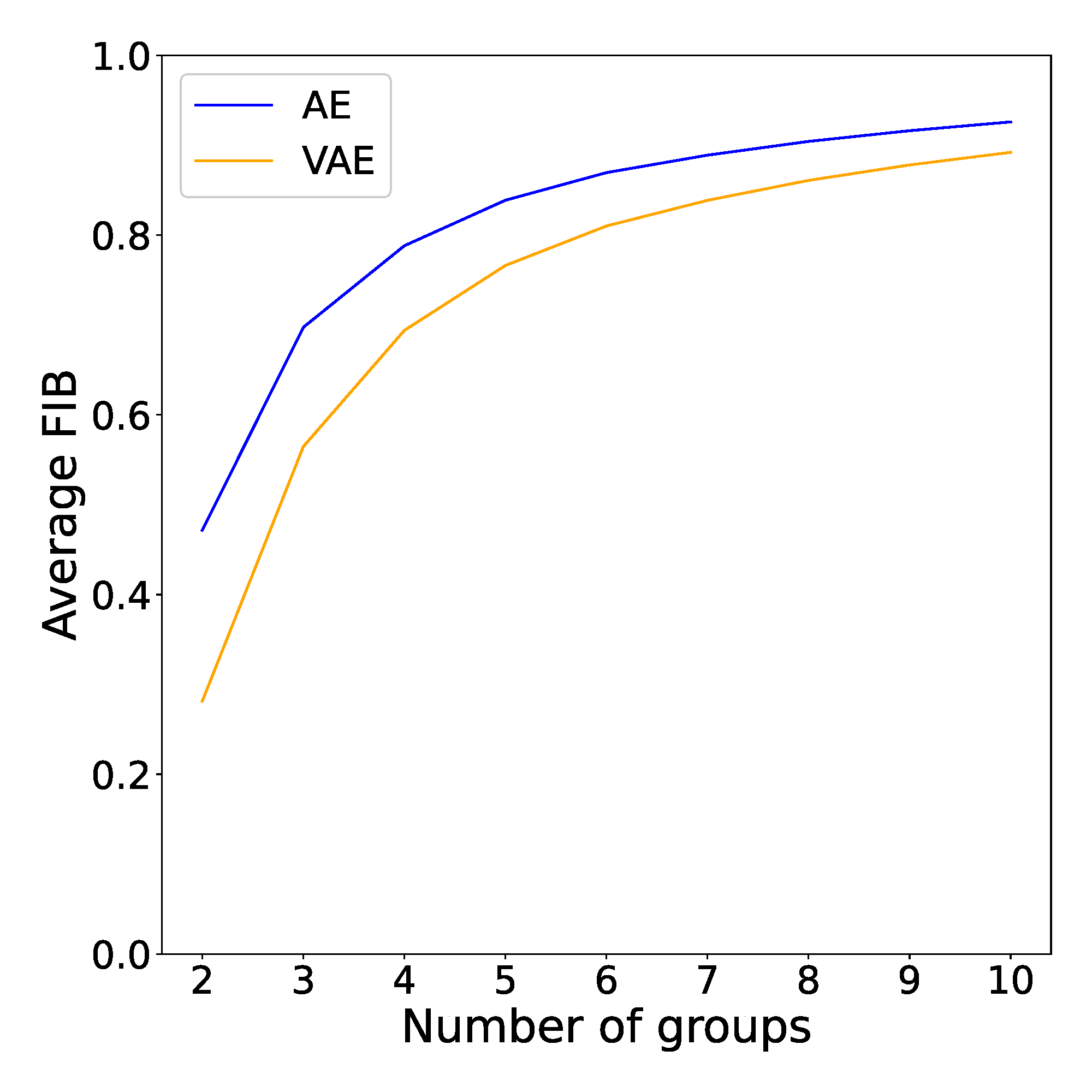}
            \caption{MNIST}
            \label{fig:exp_vae_mnist}
            \end{subfigure}
        \caption{We compare the FIB scores for AutoEncoders and Variational AutoEncoders for different numbers of groups of features on SARCOS and MNIST. \ref{fig:exp_vae_sarcos} Using an AutoEncoder with latent dimension 4,  \ref{fig:exp_vae_mnist} Using a Variational AutoEncoder with latent dimension 4 for SARCOS, and with latent dimension 256 for MNIST.} 
        \label{fig:exp_ae_vae_sarcos_mnist}
    \end{figure}

    We compare FIB scores obtained using AutoEncoders and Variational AutoEncoders on the SARCOS and MNIST datasets. Table \ref{tab:ae_vs_vae_sarcos_mnist} shows that FIB scores do not drastically change between AE and VAE if we consider all features. We considered the results after training 100 models for 1000 epochs on SARCOS \ref{fig:exp_vae_sarcos} and 300 for MNIST \ref{fig:exp_vae_mnist}. For the SARCOS dataset, the sizes of the layers of the AutoEncoders are 21,16 and 4. For MNIST the sizes are 784, 512, 256.
    We see that both architectures yield high a FIB score in Table \ref{tab:ae_vs_vae_sarcos_mnist}. We further compare performance based on the number of groups used, and in the case of SARCOS the VAE tends to have slightly higher, 0.004 for 2 groups and 0.003 for 3 groups, while for MNIST the VAE drops by 0.291 points and 0.132 for 2 and 3 groups FIB respectively. 
    \begin{table}
        \caption{FIB and grouped FIB scores obtained at the final epoch on AutoEncoders and Variational AutoEncoders for SARCOS and MNIST}
        \label{tab:ae_vs_vae_sarcos_mnist}
        \centering
        \begin{tabular}{c c c c c} 
        \toprule
         Dataset & SARCOS & SARCOS & MNIST & MNIST \\
         Type & AE & VAE & AE & VAE \\
         \midrule
         FIB & $0.997 \pm 0.000$ & $0.997 \pm 0.004$ &  $0.999 \pm 0.000$ & $0.999 \pm 0.000$ \\
         FIB 2 & $0.965 \pm 0.006$ & $0.969 \pm 0.002$ & $0.472 \pm 0.011$ & $0.281 \pm 0.003$ \\
         FIB 3 & $0.975 \pm 0.004$ & $0.978 \pm 0.001$ & $0.697 \pm 0.008$ & $0.565 \pm 0.002$ \\
             \bottomrule
        \end{tabular}
    \end{table}

   \subsection{Model Selection and FIB}
    \label{sec:exp_model_selection}
    For each of the 100 models trained in section \ref{sec:exp_datasets}: We use the weights associated with the best performing epoch with respect to the loss over the validation set. Using these versions of our models, we use the encoder section to extract representations from our train, validation and test splits.
    
\begin{figure}[h]
        \centering
        \begin{subfigure}{0.4\textwidth}
            \centering
            \includegraphics[width=\textwidth]{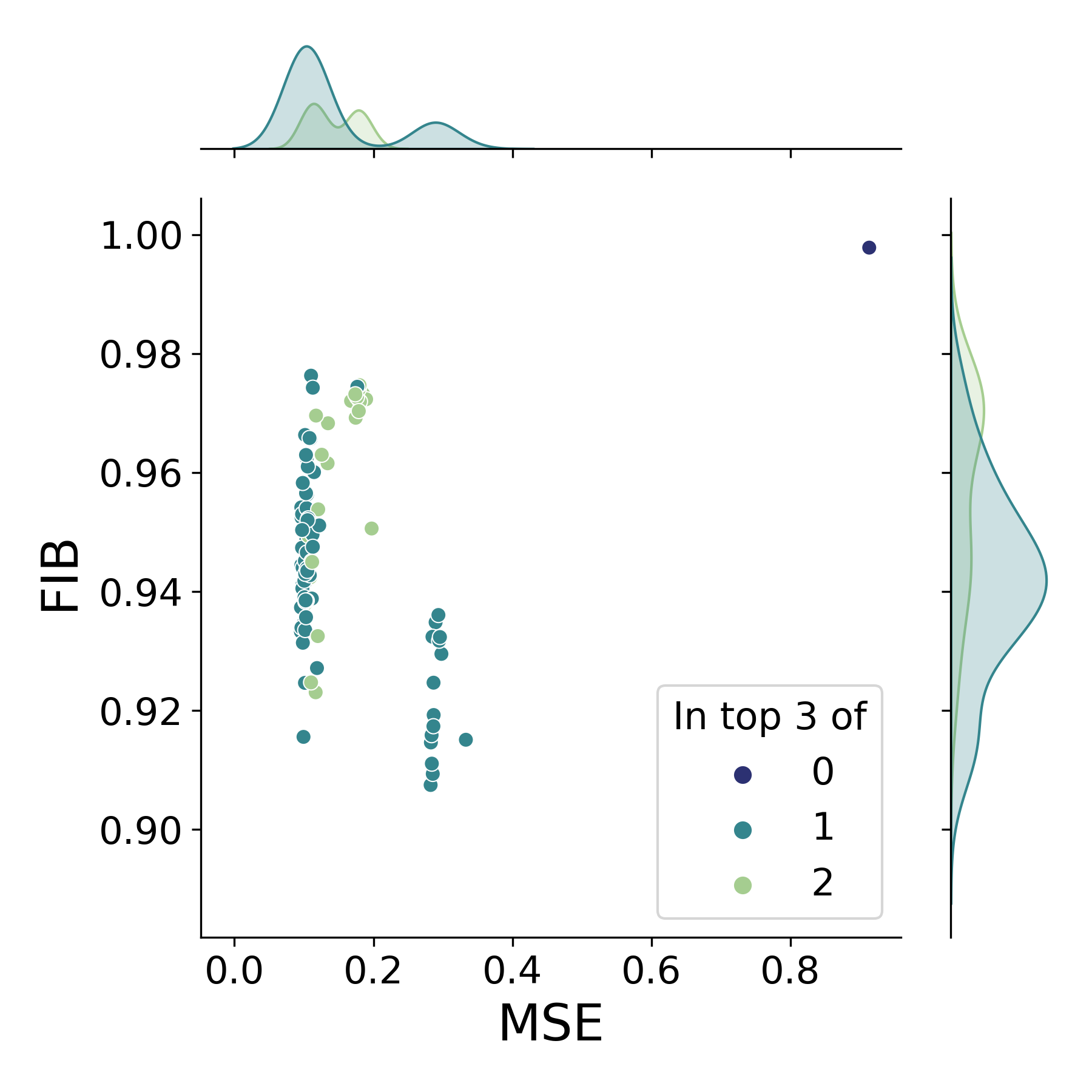}
            \caption{Iris}
            \label{fig:model_iris}
        \end{subfigure}
        \begin{subfigure}{0.4\textwidth}
            \centering
            \includegraphics[width=\textwidth]{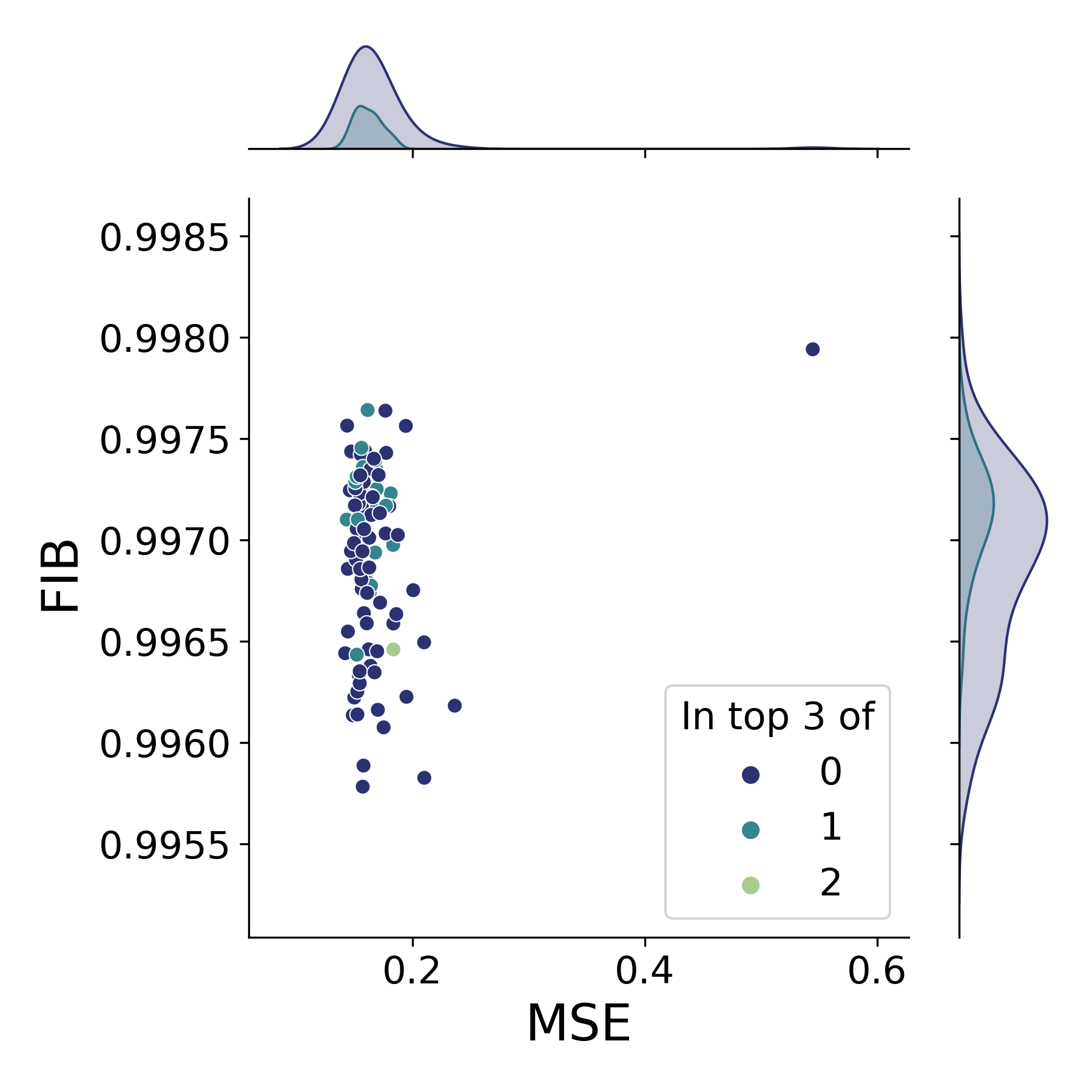}
            \caption{SARCOS}
            \label{fig:model_sarcos}
        \end{subfigure}
    \caption{For each dataset we ranked the AutoEncoders based on the performance of the subsequent One Versus Rest classifications for Iris and regressions outputs for SARCOS. Each is colored based on the number of tasks in which it ranks among the top 3.}
    \label{fig:models_iris_and_sarcos}
    \end{figure}

    After extracting the representations, we train 3 Logistic Regression for Iris, 7 Linear Regression for SARCOS to predict each dataset's respective property of interest. For instance, on Iris each Logistic Regression predicts a single class among "Setosa", "Versicolour" and "Virginica". For Iris, we consider the AutoEncoders with sizes 4, 2, for SARCOS we consider the AutoEncoders with sizes 21, 16, 4.
    The performances reported here are obtained on the test set.
    Figure \ref{fig:models_iris_and_sarcos}, shows the performance obtained by the models using the learned representations. In the case of Iris, no model was able to rank in the top 3 of all 3 models at once. In the case of SARCOS no models were able to rank in the top 3 of 4 to 7 models. In average models that ranked in the top 3 on most classifiers (resp. regressors) have a higher FIB score.
    
    Therefore, representations from models with higher FIB score are better suited to cover more downstream tasks.
    
\section{Limitations and Restrictions}
\label{section:limitations_and_restrictions}
    Depending on the function used to compute the Balance Error, it is possible to have the same value of FIB for different combinations of imbalance. Nonetheless, using the Feature Impact vector from equation \ref{equation:feature_impact_vector}, we can obtain a fine grain overview of the contributions of each feature.
    
    Furthermore, the FIB score only gives information about the balance between the impact of the features in the error and does not provide information about the intensity of said error. Therefore, the FIB score should be used in combination with other metrics to assess the performance of models.
    
    We only considered Fully Connected AutoEncoders in this work, further work could be done to analyze the impact of using AutoEncoders using Convolutional Layers.

\section{Conclusion and Future work}
    \label{section:conclusion_and_future_work}
    We introduced a method to quantify the balance of the contribution of features to an error between two vectors, called Feature Impact Balance (FIB) score. FIB relies on an Internal Error function to evaluate the error between two vectors, then use a Balance Error function to quantify how balanced the contributions of each feature are with respect to the errors. The FIB score identifies whether a single feature or a group of features contributes to most errors. We demonstrated that the FIB score can be adapted to different configurations to assess imbalance within the contribution of individual features or groups of features in the errors. We show that AutoEncoders tend to balance the impact of the features during the training phase. We observe that representations obtained from models with higher FIB scores tend to be beneficial for multiple tasks.
    
    \paragraph{Future Work.}
        \begin{itemize}
            \item \textit{Training Using FIB.} It could be possible to train models using the FIB score. One way would be to add it as a penalization term. Nevertheless, as it is possible to achieve high FIB with high reconstruction error in the case of AutoEncoders, optimizing for the FIB score might not lead to a stable learning, one way to overcome this issue could be to increase the importance of the loss due to the FIB score after some epoch, or after lowering the reconstruction loss first.
 
            \item \textit{FIB and Multimodal Machine Learning.} Feature Grouping could be used to study imbalance between modalities. If we were to consider  the multimodal representation problems \cite{baltrusaitisMultimodalMachineLearning2019}, we could monitor which models spread their error contribution equally over multiple modalities.
   
            \item \textit{Different Data types.} We could leverage the Internal Error function to obtain FIB score based on different data types. If we are able to quantify the error produced by a specific feature, we should be able to compute the impact of each feature on the total error and obtain a FIB score.
            
             \item \textit{FIB and Difference between clusters.} FIB score could be computed over averaged features from a specific class in the case of classification, for instance we could look at the difference between the average vectors of class A versus the average vector of another class B.
        \end{itemize}

\begin{ack}
The authors would like to thank Ranya Aloufi, Anastasia Borovykh, Alain Miranville, Yuchen Zhao
for numerous comments, technical questions, references, and invaluable suggestions for presentation
that led to an improved text.

Xavier F. Cadet is supported by UK Research and Innovation [UKRI Centre for Doctoral Training in AI for Healthcare grant number EP/S023283/1].
\end{ack}

\bibliography{references}
\bibliographystyle{style/icml2022}

\section{Appendix}
\appendix

\section{Training Methodology}
    \paragraph{Dataset Information.}
    We conducted our experiments on 3 datasets:
    \begin{itemize}
        \item Iris: We use the version provided by Scikit-Learn. We apply z-score normalization after splitting the data.
        \item SARCOS: Using the dataset from the companion website of \cite{rasmussenGaussianProcessesMachine2019}. We only use the training set, as the original test sets overlaps with the original training set. We apply Z-score normalization after splitting the dataset.
        \item MNIST: Using the version provided by the torchvision package. We merged both the train and test sets into a single one splitting. We do not preprocess the data otherwise.
    \end{itemize}
    We split each of the dataset into 3 sets, a test set 20\%, 24\% validation, 56\% training set.
    We use the training set and validation to train the AutoEncoders and Variational AutoEncoders (Sections \ref{sec:exp_datasets}, \ref{sec:exp_aes}, \ref{sec:exp_feature_grouping}).
    We use all three sets when considering Classifications and Regression tasks using the learned representations (Section \ref{sec:exp_model_selection}
    
    \paragraph{Training Details.}
    Models were trained on an internal cluster with various computing power allocations. 
    We train our models using ADAM optimizer with learning rate 1e-3 for AutoEncoders and 1e-4 for the Variational AutoEncoders.
    We use an internal cluster composed of Nvidia RTX 6000 GPUs.
    For SARCOS, one model trained for 1000 epochs takes in average 1h-2h, leading to 100h-200h for a given architecture.
    
\section{Theorems and their proofs}
    \label{appendix:theorems_and_proofs}
    \paragraph{Theorem \ref{theorem:reach_max}}
    Let $f:S \rightarrow \mathbb{R}$ be continuous and convex, where $S \subset \mathbb{R}^{M}$ is the $M$-simplex defined by the points $a_1$, $\cdot \cdot \cdot $, $a_{M+1}$, $M\ge 1$. Then $f$ reaches its maximum at one of the $a_i$'s.

    \begin{proof}
    First note that $f$ is continuous on the nonempty compact set $S$. It thus reaches its maximum.
    
    Let then $a\in S$. We can thus write
    
    $$a=\sum _{i=1}^{M+1}\lambda _ia_i,$$
    
    where $\lambda _i\in [0,1]$, $i=1$, $\cdot \cdot \cdot $, $M+1$, $\sum _{i=1}^{M+1}\lambda _i=1$. Since $f$ is convex, we have
    
    $$f(a)=f(\sum _{i=1}^{M+1}\lambda _ia_i)\le
    \sum _{i=1}^{M+1}\lambda _if(a_i)$$
    $$\le \max (f(a_1),\ \cdot \cdot \cdot ,f(a_{M+1})),$$
    
    which proves the theorem.
\end{proof}

    \paragraph{Theorem \ref{theorem:maximal_value}}
    Let $A = \{(c_1, ..., c_M) \in \mathbb{R}^M, c_i \in [0, 1], i = 1, .., M, \sum_{i=1}^{M}c_i = 1\}$. We set
$\varphi (c_1, ..., c_M) = \sum_{i= 1}^{M}|c_i - \frac{1}{M}|$ and $\psi (c_1, ..., c_M) = \sum_{i = 1}^{M}(c_i - \frac{1}{M})^2$. Then, $\max_{A} \varphi = 2\frac{ M -1}{M}$ and $\max_{A} \psi = \frac{M - 1}{M}$.
    \begin{proof}
    For $M=1$ we have $c_1=1$, so that both $\varphi $ and $\psi $ vanish and the assertion of the theorem holds true.
    
    Let us now assume that $M\ge 2$. Note that $\varphi $ and $\psi $ are continuous and convex (as sums of continuous and convex functions). Then note that we wish to maximize $\varphi $ and $\psi $ on the $(M-1)$-simplex defined by the points (in $\mathbb{R}^{M+1}$) $(1,0,\cdot \cdot \cdot ,0)$, $\cdot \cdot \cdot $, $(0,0,\cdot \cdot \cdot ,0,1)$ in the affine hyperplane $\sum_{i=1}^{M}c_i = 1$.
    The result then follows from Theorem \ref{theorem:maximal_value}.
\end{proof}

\section{Example of Feature Grouping Algorithm}
    \label{appendix:feature_grouping_algorithm}
    \section{Feature Grouping}
\begin{figure}[ht]
    \vskip 0.2in
        \begin{center}
            \centerline{\includegraphics[width=\columnwidth]{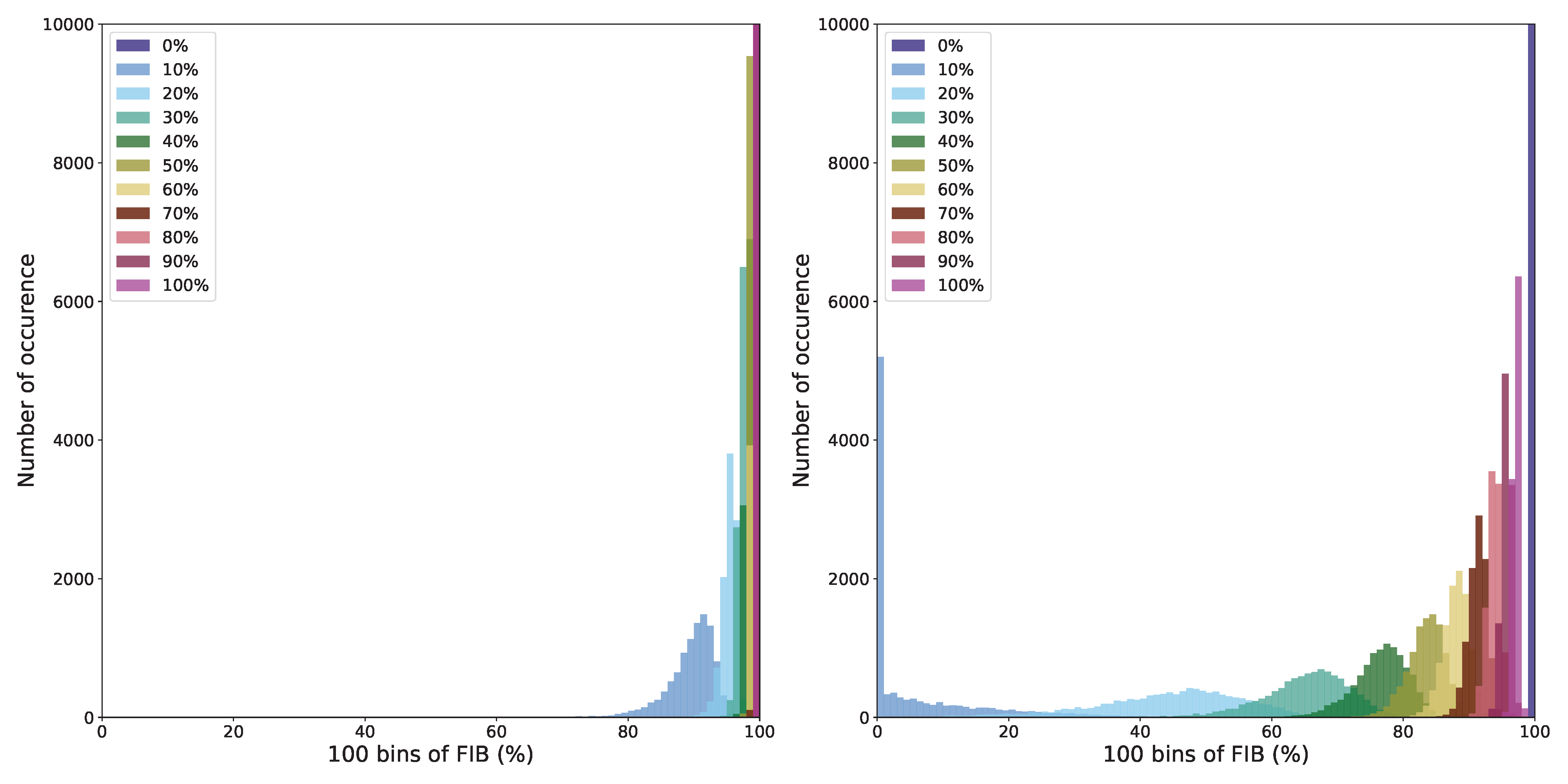}}
            \caption{
            Uniform noise injection in $128$ dimensional vectors using (left) FIB and (right) Grouped FIB with $10$ groups. Colors are associated with \% of dimensions with noise injection. After grouping, varying percentages of perturbation lead to different FIB scores.}
            \label{figure:supp_128}
        \end{center}
    \vskip -0.2in
\end{figure}

\begin{figure}[ht]
    \vskip 0.2in
        \begin{center}
        
            \centerline{\includegraphics[width=\columnwidth]{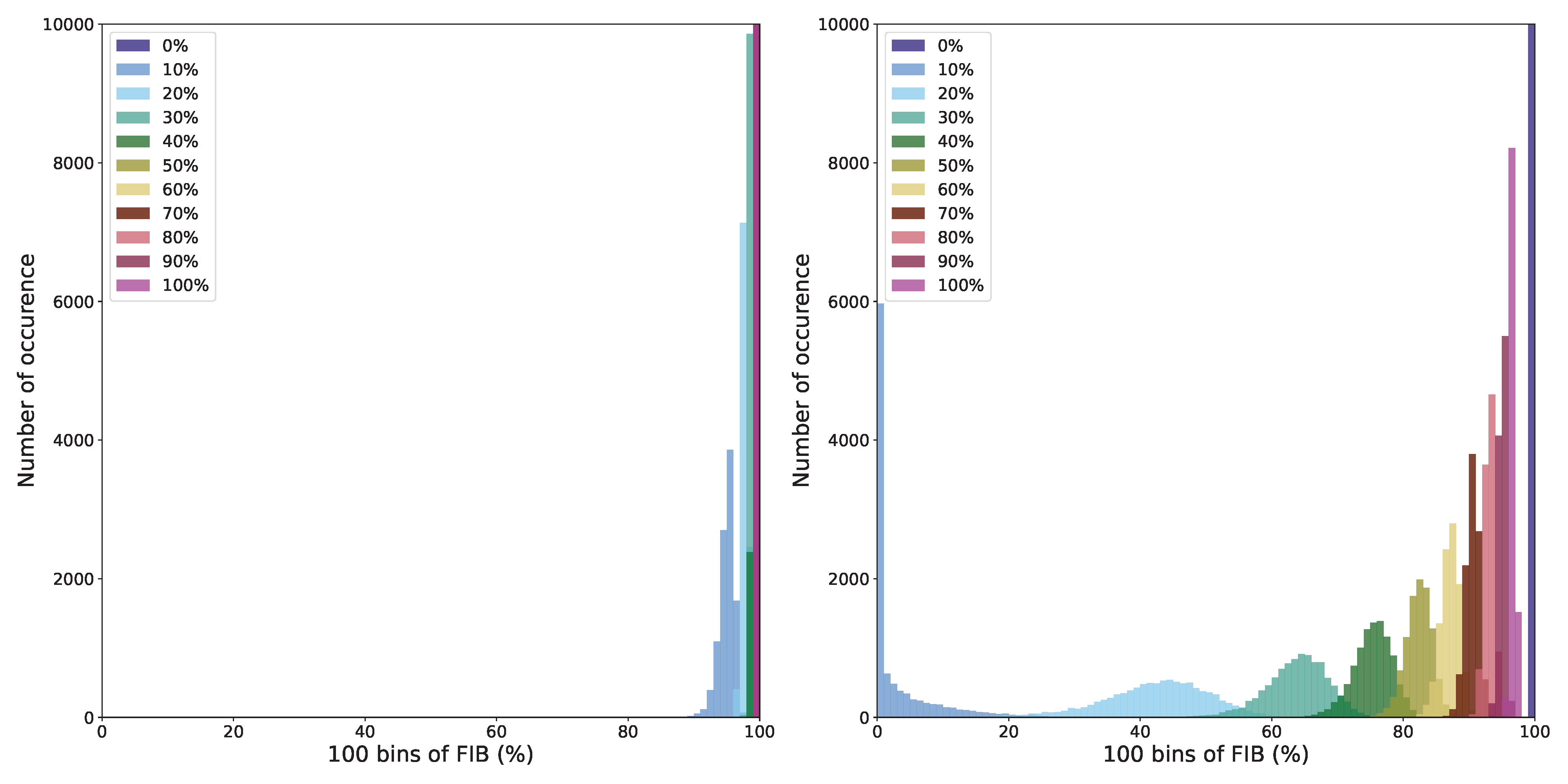}}
            \caption{
            Uniform noise injection in $256$ dimensional vectors using (left) FIB and (right) Grouped FIB with $10$ groups. Colors are associated with \% of dimensions with noise injection. After grouping, varying percentages of perturbation lead to different FIB scores.}
            \label{figure:supp_256}
        \end{center}
    \vskip -0.2in
\end{figure}

\begin{figure}[ht]
    \vskip 0.2in
        \begin{center}
            \centerline{\includegraphics[width=\columnwidth]{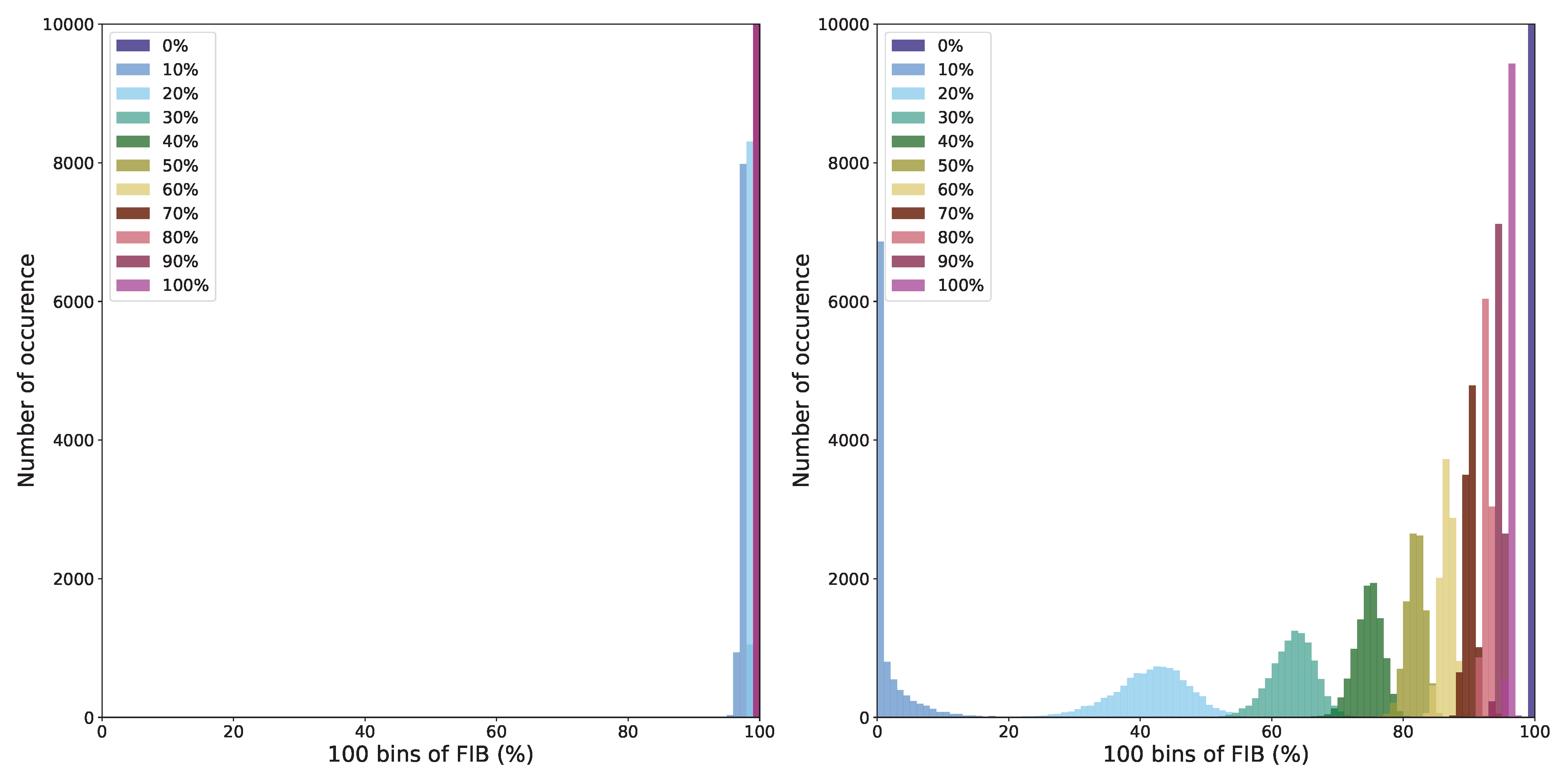}}
            \caption{
            Uniform noise injection in $512$ dimensional vectors using (left) FIB and (right) Grouped FIB with $10$ groups. Colors are associated with \% of dimensions with noise injection. After grouping, varying percentages of perturbation lead to different FIB scores.}
            \label{figure:supp_512}
        \end{center}
    \vskip -0.2in
\end{figure}

\begin{figure}[ht]
    \vskip 0.2in
        \begin{center}
            \centerline{\includegraphics[width=\columnwidth]{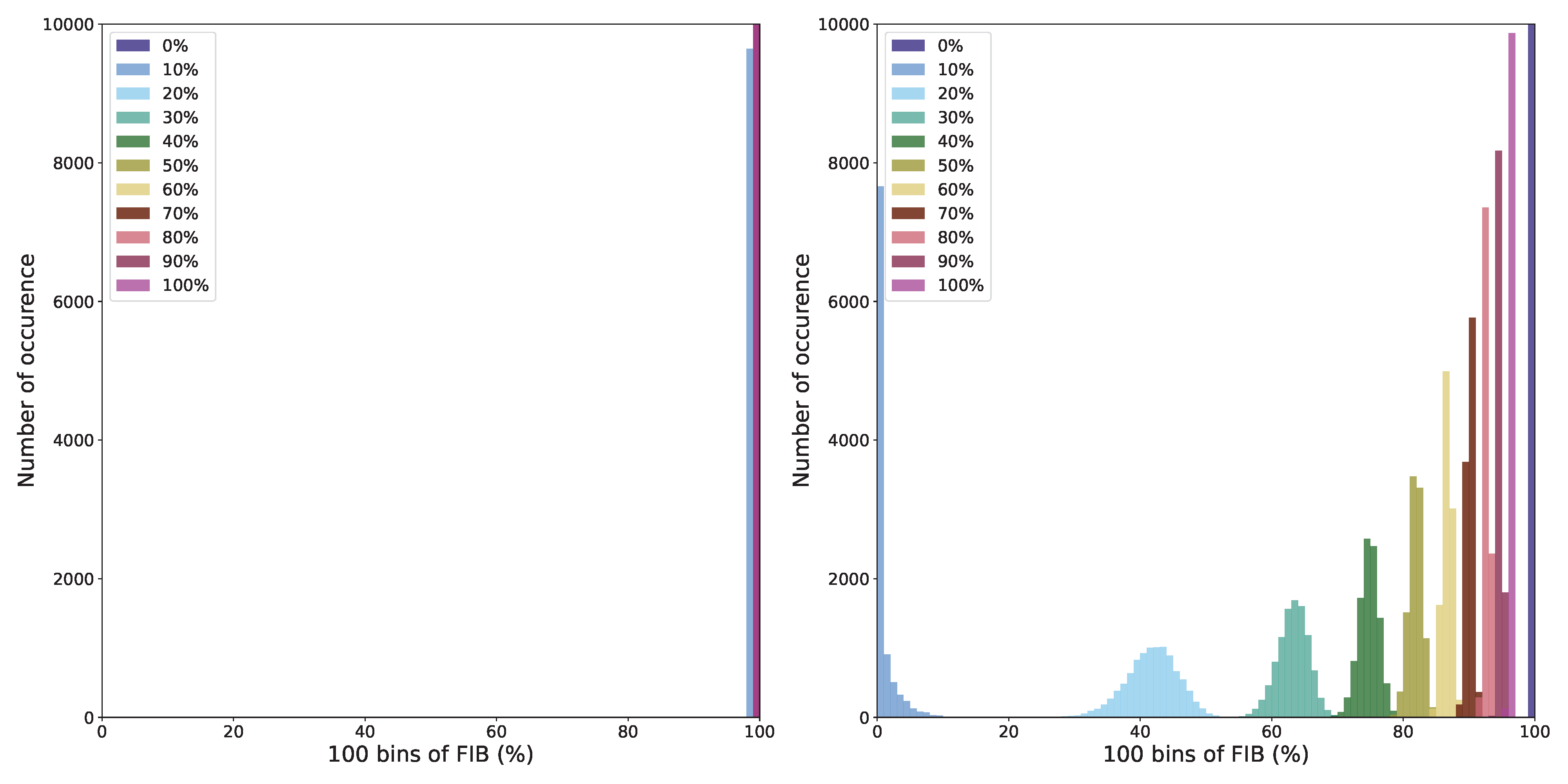}}
            \caption{
            Uniform noise injection in $1024$ dimensional vectors using (left) FIB and (right) Grouped FIB with $10$ groups. Colors are associated with \% of dimensions with noise injection. After grouping, varying percentages of perturbation lead to different FIB scores.}
            \label{figure:supp_1024}
        \end{center}
    \vskip -0.2in
\end{figure}

Figures \ref{figure:supp_128}, \ref{figure:supp_256}, \ref{figure:supp_512}, \ref{figure:supp_1024} showcase the difference between FIB scores and FIB scores obtained using Feature Grouping.
From top to bottom, the figures represent the effects of FIB and Grouped FIB over $128, 256, 512$ and $1024$ Features.
As the number of features grows, the noise injection packs the FIB (right) score closer to the maximal value. Nonetheless, using Feature Grouping leads to FIB scores spreading from the minimal score to the maximal score.
The values are reported based on results from $10000$ noise injections on the synthetic data discussed in section 4.3.

\end{document}